\documentclass[10pt,twocolumn,letterpaper]{article}

\usepackage{cvpr}
\usepackage{times}
\usepackage{epsfig}
\usepackage{graphicx}
\usepackage{subfigure}
\usepackage{amsmath}
\usepackage{amssymb}
\usepackage{comment}

\usepackage[breaklinks=true,bookmarks=false]{hyperref}

\cvprfinalcopy % *** Uncomment this line for the final submission

\def\cvprPaperID{****} % *** Enter the CVPR Paper ID here

\begin{document}

%%%%%%%%% TITLE
\title{Collaborative Descriptors:\\Convolutional Maps for Preprocessing}

\author{
Hirokatsu Kataoka, Yutaka Satoh \\
AIST\\
%Tsukuba, Ibaraki, Japan\\
{\tt\small \{hirokatsu.kataoka, yu.satou\}@aist.go.jp}
\and
Kaori Abe\\
AIST, Tokyo Denki University\\
%Tsukuba, Ibaraki, Japan\\
%Adachi, Tokyo, Japan\\
{\tt\small abe.keroko@aist.go.jp}
\and
Akio Nakamura\\
Tokyo Denki University\\
%Adachi, Tokyo, Japan\\
{\tt\small nkmr-a@cck.dendai.ac.jp}
}

\maketitle
%\thispagestyle{empty}

%%%%%%%%% ABSTRACT
\begin{abstract}
   The paper presents a novel concept for collaborative descriptors between deeply learned and hand-crafted features. To achieve this concept, we apply convolutional maps for pre-processing, namely the convovlutional maps are used as input of hand-crafted features. We recorded an increase in the performance rate of +17.06\% (multi-class object recognition) and +24.71\% (car detection) from grayscale input to convolutional maps. Although the framework is straight-forward, the concept should be inherited for an improved representation. 
\end{abstract}

%%%%%%%%% BODY TEXT
\section{Introduction}

Neural networks computationally mirror the architecture of the human brain. Recent architectures have been made deeper in order to provide higher-level representations of object. Especially in computer vision, deep convolutional neural networks (DCNN) are becoming increasingly widely used for image recognition and object detection. Representative deep models such as VGGNet~\cite{SimonyanICLR2015} have been proposed in the field of computer vision. 

\begin{figure}[t]
\begin{center}
   \includegraphics[width=0.9\linewidth]{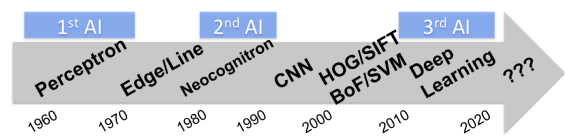}
\end{center}
   \caption{The $1^{st}$, $2^{nd}$ and $3^{rd}$ AI booms and winters: ($1^{st}$ boom) perceptron, ($1^{st}$ winter) edge/line detection, ($2^{nd}$ boom) neocognitron, ($2^{nd}$ winter) HOG/SIFT/BoF/SVM, ($3^{rd}$ boom [now]) deep learning. Here we anticipate a next framework after $3^{rd}$ AI.}
\label{fig:ai}
\end{figure}

The use of neural net and hand-crafted feature is repeated in the computer vision field (see Figure~\ref{fig:ai}). The suggestive knowledge motivates us to study an advanced hand-crafted feature after the DCNN in the recent $3^{rd}$ AI. We anticipate that the hand-crafted feature should collaborate with deeply learned parameters since an outstanding performance is achieved with the automatic feature learning.

In the paper, we present a novel concept for collaborative descriptors which jointly work both DCNN and hand-crafted feature. We apply convolutional maps for pre-processing as a first implementation. The framework is simple, but it is effective for recognition tasks. We directly access the convolutional maps in order to improve the performance rate  of hand-crafted features. We try to evaluate several hand-crafted features such as scale invariant feature transform (SIFT) and bag-of-words (BoW)~\cite{CsurkaECCVW2004} (object recognition) and higher-order local autocorrelation (HLAC)~\cite{OtsuIAPR1988} (car detection). Then, the recognition rates for all hand-crafted features are compared using convolutional maps and grayscale images that have the same settings for the image description.

\section{Convolutional maps for preprocessing}

\begin{figure}[t]
\begin{center}
   \includegraphics[width=1.0\linewidth]{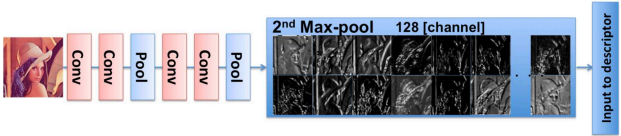}
\end{center}
   \caption{Framework for the proposed method}
\label{fig:framework}
\end{figure}

Here, we introduce a framework for combining convolutional maps and hand-crafted features. Figure~\ref{fig:framework} shows the process for using convolutional maps as a preprocessor. We employ hand-crafted features, namely SIFT+BoW~\cite{CsurkaECCVW2004} and HLAC~\cite{OtsuIAPR1988}.

Given a convolutional map $M_{k}$ with VGG~\cite{SimonyanICLR2015} for kernel $k$, our goal is to extract a feature vector $V$ from the map and define a classification label $L$ as a function of SVM, $L = f_{svm}(V)$. We use VGGNet to create convolutional maps. 

%VGGNet is a simple and well-organized architecture. We employ a 16-layer model in order to access the convolutional maps in VGGNet. VGGNet convolves the maps using a patch size of $3\times3$ [pixels]. This small and deeply iterative patch size results in a high degree of nonlinearity in the feature space. Although the architecture is based on AlexNet, the deeper structure is more nonlinear and provides better performance. 
The $2^{nd}$ max-pooling layer is used as a convolutional map that contains the $56\times56\times128$ feature map. We can handle 128 maps with dimensions of $56\times56$ [pixel]. The patch and kernel sizes are suitable for balanced feature extraction. 

We can obtain a convolutional map $M^{conv}$ and max-pooling map $M^{mp}$ for kernel $k$ from input map $M$ as below:
\begin{eqnarray}
  M^{conv}_{k} &=& {\displaystyle \sum_{x}^{X}\sum_{y}^{Y}} f_{kxy}M_{xy} + b_{kxy} \\
  M^{mp}_{k} &=& \max_{ij} M^{conv}_{ijk}
\end{eqnarray}
where $f$ is a filter and $b$ is a bias in CNN. $x, y, i, j$ are the indices of the two-dimensional map. Hereafter, the $2^{nd}$ max-pooling layer $M^{mp}$ is the input image.

\section{Results}

\begin{figure}[t]
\begin{center}
   \includegraphics[width=0.7\linewidth]{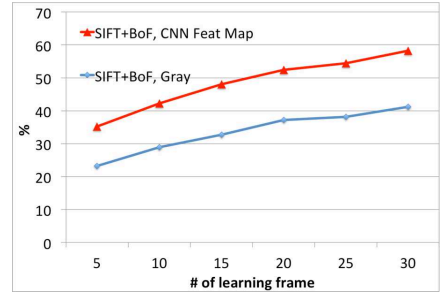}
\end{center}
   \caption{Recognition rate for the Caltech 101 dataset}
\label{fig:caltech}
\end{figure}

Comparisons of the hand-crafted features from convolutional maps and grayscale images are shown in Figure~\ref{fig:caltech} and Table~\ref{tab:uiuccars}, respectively. We used a support vector machine (SVM) as a two- or multi-class classifier.

SIFT+BoW was employed for the Caltech 101 dataset~\cite{FeiFeiCVPR2004}, since it is a kind of object categorization. According to Figure~\ref{fig:caltech}, the convolutional maps allow us to clearly distinguish between object classes. In the case of SIFT+BoW with convolutional maps, the codeword dictionary was shared across all 128 kernels. The dimension of the vector was 1,000, even when using convolutional maps. SIFT+BoW with convolutional maps (rate: 58.28\%) was +17.06\% better than the SIFT+BoW for the grayscale image (rate: 41.22\%) on the Caltech 101 dataset. The results show that the use of preprocessing gives a perspective for the learning-based preprocessing with CNN. Although the CNN architecture is based on the ImageNet pre-trained model, a more sophisticated model such as fine tuning with objective data may improve preprocessing.

HLAC was applied to the UIUC cars dataset~\cite{AgarwalTPAMI2004}. The HLAC setting in convolutional maps was used to extract 25 dimension features per learned kernel, therefore we can extract a 25 [dim] $\times$ 128 [kernel] = 3,200 [dim] HLAC feature from an image. Table~\ref{tab:uiuccars} shows the proposed setting (98.82\%) performed better than the conventional grayscale setting (74.11\%). We get a +24.71\% improvement through using the convolutional maps. The preprocessing of HLAC usually requires binarization; however, we can extend this to a combination of convolution and binarization with convolutional maps.

The convolutional maps allow us to improve the expressiveness of the feature vector. The next step will make use of convolutional maps with well-organized approaches.

%\begin{figure}[t]
%\begin{center}
%   \includegraphics[width=0.9\linewidth]{figure/uiuccar.eps}
%\end{center}
%   \caption{Recognition rate for the UIUC cars dataset}
%\label{fig:uiuccar}
%\end{figure}

\begin{table}[t]
\begin{center}
\caption{Recognition rate for the UIUC cars dataset}
\begin{tabular}{lc}
Approach & \% \\
\hline
HLAC, Grayscale & 74.11 \\
HLAC, ConvMaps & 98.82 \\
\hline
\end{tabular}
\label{tab:uiuccars}
\end{center}
\end{table}

\section{Conclusion}

The paper presented a novel concept of collaborative descriptors which combine deep convolutional neural network (DCNN) into hand-crafted features. We assigned convolutional maps as a preprocessing. We recorded an increase in the performance rate of +17.06\% and +24.71\% for the Caltech101 and UIUC cars datasets, respectively. 

In the future, we must adjust the parameters of the convolutional maps, hand-crafted features and their integration. This paper only considered old-fashioned features and datasets; however, the use of convolutional maps will likely result in even greater improvements in the performance rate when used in conjunction with more sophisticated features.

{\small
\bibliographystyle{ieee}
\bibliography{convmap}

\begin{thebibliography}{1}\itemsep=-1pt

\bibitem{AgarwalTPAMI2004}
S.~Agarwal, A.~Awan, and D.~Roth.
\newblock Learning to detect objects in images via a sparse, part-based
  representation.
\newblock {\em IEEE Trans. on Pattern Analysis and Machine Intelligence
  (TPAMI)}, 286(11), 2004.

\bibitem{CsurkaECCVW2004}
G.~Csurka, C.~R. Dance, L.~Fan, J.~Willamowski, and C.~Bray.
\newblock Visual categorization with bags of keypoints.
\newblock ECCVW, 2004.

\bibitem{FeiFeiCVPR2004}
L.~Fei-Fei, R.~Fergus, and P.~Perona.
\newblock Learning generative visual models from few training examples: an
  incremental bayesian approach tested on 101 object categories.
\newblock CVPR, 2004.

\bibitem{OtsuIAPR1988}
N.~Otsu and T.~Kurita.
\newblock A new scheme for practical flexible and intelligent vision systems.
\newblock IAPR Workshop on Computer Vision, 1988.

\bibitem{SimonyanICLR2015}
K.~Simonyan and A.~Zisserman.
\newblock Very deep convolutional networks for large-scale image recognition.
\newblock International Conference on Learning Representation (ICLR), 2015.

\end{thebibliography}
}

\end{document}